\documentclass[letterpaper, 10 pt, conference]{ieeeconf}
\usepackage[lcgreekalpha,notext]{stix}
\usepackage{siunitx}

\usepackage{graphics} % for pdf, bitmapped graphics files
\usepackage{epsfig} % for postscript graphics files
\usepackage{graphicx}

\usepackage{amsmath}

\usepackage{algorithmic}
\usepackage[ruled,linesnumbered]{algorithm2e}

\usepackage{url}
\usepackage{textcomp}
\usepackage{hyperref}
\usepackage{tabularx}
\usepackage{tabulary}
\usepackage{xfrac,nicefrac}
\usepackage{makecell}

\usepackage{multirow}
\newcolumntype{Y}{>{\centering\arraybackslash}X}
\usepackage{threeparttable}

\usepackage{colortbl}
\usepackage{xcolor}
\definecolor{sd}{gray}{0.5}
\definecolor{myblue}{rgb}{0.239215686,0.239215686,0.6}
\definecolor{mygreen}{rgb}{0.21568628,0.6901961,0.21568628}
\definecolor{myred}{rgb}{0.909803922,0.082352941,0.082352941}
\definecolor{mygray}{gray}{.9}
\definecolor{mypink}{rgb}{.99,.91,.95}
\usepackage{listings}

\usepackage{footnote}

\usepackage{xspace}
\newcommand{\etal}[0]{{\em et al.~}}
\newcommand{\eg}[0]{{\em e.g.,~}}
\newcommand{\ie}[0]{{\em i.e.,~}}

\usepackage{eurosym}
\usepackage{booktabs, tabularx}

\usepackage{tikz}
\usetikzlibrary{calc, math}
\makeatletter
\let\NAT@parse\undefined
\makeatother
\usepackage[numbers]{natbib}

\usepackage{afterpage}
\providecommand{\U}[1]{\protect\rule{.1in}{.1in}}   
\IEEEoverridecommandlockouts
\begin{document}
\title{\LARGE \bf Maintaining a Reliable World Model \\using Action-aware Perceptual Anchoring}
% \title{\LARGE \bf Improving Perceptual Anchoring using Agent Actions}
% \title{\LARGE \bf Object Permanence with Action Anchoring for World State Representation}
% "reliable world model tracking" using rule-based approach in CA

\author{Ying Siu Liang$^1$, Dongkyu Choi$^1$, Kenneth Kwok$^1$
% \author{Anonymous
\thanks{$^1$Social and Cognitive Computing, Institute of High Performance Computing, Agency for Science, Technology and Research, Singapore.}
\thanks{Email: \{liangys, choi\_dongkyu, kenkwok\}@ihpc.a-star.edu.sg}
}

\maketitle

\vskip -0.15in
%%%%%%%%%%%%%%%%%%%%%%%%%%%%%%%%%%%%%%%%%%%%%%%%%%%%%%
\begin{abstract}
Reliable perception is essential for robots that interact with the world. But sensors alone are often insufficient to provide this capability, and they are prone to errors due to various conditions in the environment. Furthermore, there is a need for robots to maintain a model of its surroundings even when objects go out of view and are no longer visible. This requires anchoring perceptual information onto symbols that represent the objects in the environment. In this paper, we present a model for action-aware perceptual anchoring that enables robots to track objects in a persistent manner. Our rule-based approach considers inductive biases to perform high-level reasoning over the results from low-level object detection, and it improves the robot's perceptual capability for complex tasks. We evaluate our model against existing baseline models for object permanence and show that it outperforms these on a snitch localisation task using a dataset of 1,371 videos. We also integrate our action-aware perceptual anchoring in the context of a cognitive architecture and demonstrate its benefits in a realistic gearbox assembly task on a Universal Robot.
\end{abstract}

%%%%%%%%%%%%%%%%%%%%%%%%%%%%%%%%%%%%%%%%%%%%%%%%%%%%%%
\section{Introduction}
\label{sec:intro}
\begin{figure}[tb]
	\centering	
 	\includegraphics[width=0.9\linewidth]{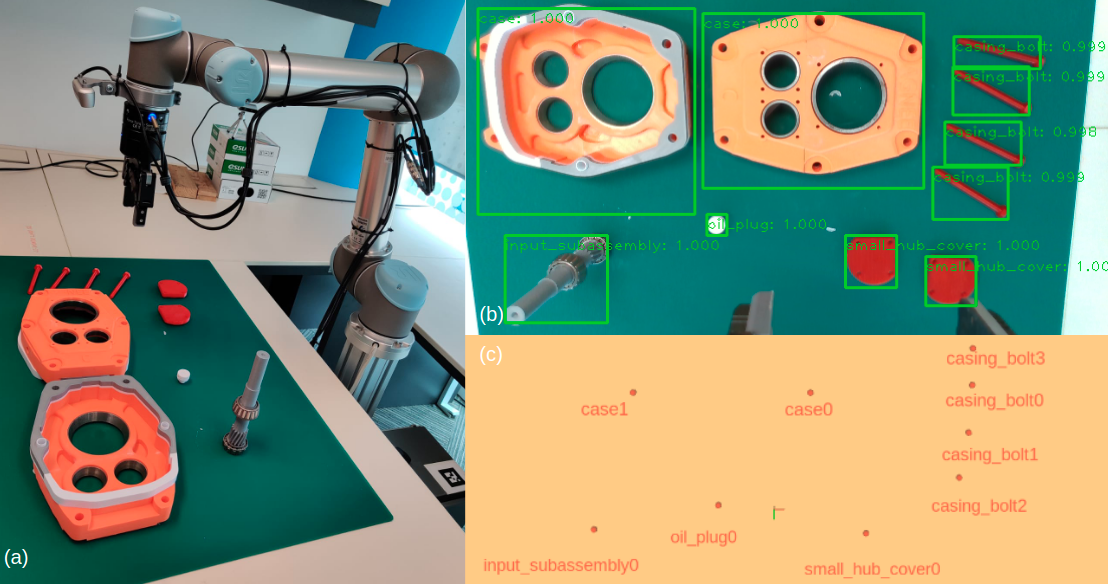}%
 	\vskip -0.11in
 	\caption{Experiment setup of the gearbox assembly task with the UR5.}
	\label{fig:setup}
	\vskip -0.15in
\end{figure}

Robots today often interact with objects and other agents in their surroundings. To perform complex tasks, robots should reason about the world before they can plan and execute actions. But the interaction among objects in the environment sometimes limits a robot's visibility, causing it to lose sight of certain objects. This is especially true when using the popular eye-in-hand setup like in our case shown in Fig.~\ref{fig:setup}. To remedy this issue, we therefore need the robot to maintain a reliable world model, with which it can estimate object states when they are no longer visible. This capability, often denoted as {\em object permanence}, is a frequently studied topic in developmental psychology that describes the ability to understand that objects continue to exist even when they become unobservable. Psychologists believe that human infants develop full object permanence at around two years of age, when they start to understand that objects still exist even if they have been completely removed from their field of view~\cite{piaget1953}.

In robotics, object permanence is an important concept with practical benefits that augments robots' searching and tracking abilities~\cite{papadourakis2010multiple,meger2008curious}, facilitates self-recognition~\cite{michel2004motion,lang2018deep}, and enables belief management and perspective taking when engaging with humans~\cite{milliez2014framework,roy2004mental}. Common approaches consider object permanence as a low-level reasoning problem that heavily relies on the visual perception capabilities % where objects are tracked based on perceived pixels, associating bounded regions of subsequent frames and comparing their colour, shape or location
\cite{papadourakis2010multiple,michel2004motion,roy2004mental}.
There are also higher-level approaches that use \textit{perceptual anchoring} to create and maintain the correspondence between symbols and perceived sensor data referring to the same physical objects~\cite{coradeschi2003introduction}. These typically use some form of high-level reasoning to track objects that disappear. For instance, there are heuristic approaches that determine objects being grasped by a hand~\cite{milliez2014framework} or track the nearest visible object instead of an unobserved one~\cite{shamsian2020learning,zhu2018distractor}. Others use probabilistic reasoning to infer occluded object states~\cite{persson2019semantic,blodow2010perception}.

But the performance of these models drops with noisy data or more complex scenarios, like when there is flickering detection of false positives or when objects are no longer detected upon insertion into another object and are subsequently {`carried'} (\eg \emph{invisible displacements}~\cite{Singer2018}).
%Shamsian \etal\cite{shamsian2020learning} address this problem in four different conditions, where objects are visible, occluded, contained (\ie covered and static), and carried (\ie covered and moving), respectively. Their deep learning architecture (OPNet) outperforms other baseline models for some complex tasks.
% While many deep learning approaches exist that address this problem, they require large amounts of data which certain applications do not have. Furthermore, neural approaches are not understandable by humans and there is little room to investigate failures in the reasoning process.
For example, when a plug is inserted into a case and is no longer visible, it becomes difficult to infer its position from vision alone, especially when the case is moved.
% The case could be moving while other objects obscure the camera view, causing us to track another object that might be closest to the plug. 
Since we know that the plug has been inserted or \textit{attached} to the case,  we can track the plug based on the position of the case. %, even after long-term occlusions.
% Thus, if an object is inserted into a container which is moved, the attachment guarantees that the object moves with it, even if it is not visible anymore.

In this paper, we define an \textit{attachment} between two objects to be the state resulting from an action that sticks one object to the other, \eg insert or screw-on, such that the former becomes a part or an extension of the latter and is physically bound to move together. To reason about attachments, the perceptual system should have knowledge about the agent actions that result in such object states. We propose a perceptual anchoring model that uses agent actions to maintain a reliable world model. 
By considering actions and their effects (\ie attachments), the robot can predict imperceivable state changes and estimate the positions of invisible objects.

To start our discussion, we first review some related work (Sec.~\ref{sec:background}). After that, we discuss the fundamental hypotheses of our work and describe our system in detail (Sec.~\ref{sec:approach}). We show that this rule-based high-level reasoning approach augments the performance of low-level approaches, by first comparing our system to baseline models on the LA-CATER dataset (Sec.~\ref{sec:experiment}) and show that our model outperforms them on a snitch localisation task.
We further demonstrate the system on a realistic robotic application with a Universal Robot (UR5) where it performs a collaborative gearbox assembly task (Sec.~\ref{sec:ur5-experiment}). 
We conclude with a discussion on future work to further improve our model (Sec.~\ref{sec:conclusion}).

\vskip -0.11in
%%%%%%%%%%%%%%%%%%%%%%%%%%%%%%%%%%%%%%%%%%%%%%%%
\section{Related Work}
\label{sec:background}

Object permanence has been addressed previously within the computer vision community, especially for multi-object tracking.
Papadourakis and Argyros~\cite{papadourakis2010multiple} handle temporally long object occlusions by assigning spatially close objects as \textit{occluders} to objects.
% Most approaches address the low-level problem of tracking clusters of detected pixels and mapping them to objects.
Lang \etal\cite{lang2018deep} trained a CNN to learn a mapping between the current sensor state and the robot's intended motor actions to predict the next sensor state. It considers joint configuration changes of its own arms that might cause object occlusions and uses the predicted image to replace the occluded part of the object in the current image.
Shamsian \etal\cite{shamsian2020learning} proposed the OPNet architecture that uses two LSTMs to track the target object's visibility, its occluder that carries it, and the target localisation.  %However, OPNet still performs significantly worse on `carried' tasks (56.04\%) compared to less complex object permanence tasks and the authors do not provide explanations for failure cases.

High-level approaches in developmental robotics have previously addressed the perceptual anchoring problem.
In Roy \etal\cite{roy2004mental}, the robot maintains a `mental model' of the environment in order to generate different viewpoints of objects. % Approaches have been fine-tuned for the conducted case study~\cite{roy2004mental}. 
In Milliez \etal\cite{milliez2014framework}, the robot makes position hypotheses on occluded objects such as the human grasping it or placing it inside another object. 
Persson \etal\cite{persson2019semantic} described a probabilistic reasoning system that infers the state of occluded objects and continue tracking through occluded movements. %to check
Blodow \etal\cite{blodow2010perception} uses probabilistic anchoring to identify objects that come back into view.
Elfring \etal\cite{elfring2013semantic} uses a Multiple Hypothesis Tracker algorithm to improve tracking reliability during occlusion and detect clutter (false positives).
Heyer and Gr\"{a}ser~\cite{heyer2012intelligent} define relative anchors to represent relations between two reference anchors and use the robot's (camera) movements to re-anchor them.
Nitti \etal\cite{nitti2014relational} implemented a relational particle filter that draws on commonsense world knowledge during tracking, such as when one object is inside another. %to check
Mininger and Laird~\cite{Mininger2019UsingDK} address anchoring errors that occur when associating perceptual data with symbolic representations and how to resolve them.
% In~\cite{meger2008curious} the robot uses object permanence to remember where the objects are that it left behind. %Not sure if needed
While the above approaches address the same problem of tracking occluded objects, only few address invisible displacements and none make use agent actions to predict attachments. We believe that attachments can significantly improve the high-level reasoning process to complement existing approaches.

% \begin{figure*}[tb]
% 	\centering	
%  	\includegraphics[width=0.95\linewidth]{figures/model-overview.png}%
%  	\caption{Pipeline of our object permanence action anchoring model (AAPA).}
% 	\label{fig:model-overview}
% \end{figure*}
\vskip -0.15in
\section{Action-aware Perceptual Anchoring}
\label{sec:approach}
We propose a perceptual anchoring model that uses agent actions together with perceptual information to maintain a world model of object anchors, described as a set of predicate symbols $\mathcal{P}=\{p_1, p_2, \dots\}$.
We believe that high-level reasoning is necessary to maintain object states that are no longer perceived by low-level object detection and can significantly improve object tracking for complex tasks such as invisible displacements. 
For this, we consider the following hypotheses that object state changes are:
\begin{enumerate}
    \item[H1.] Observable: objects do not move on their own and state changes are only caused by agent actions. Thus, we assume that object positions stay the same as last time they were seen, until they are redetected. %(required for maintaining out of view).
    \item[H2.] Gradual: objects do not `teleport', so detected objects in consecutive frames are likely to have similar attributes \eg position and size. %(required for the frame alignment process).
    \item[H3.] Consistent: objects are detected consistently over time, so they are only considered to exist/not exist in the world model if they are present/absent for multiple consecutive cycles. %This is similar to humans only being able to perceive differences at a limited framerate.
    % \item[H2.] False negatives: objs are removed if unseen over multiple cycles conf* $< 0$ (hysterisis fct) also removes false positives
\end{enumerate}

\begin{figure*}[tb]
	\centering	
 	\includegraphics[width=0.95\linewidth]{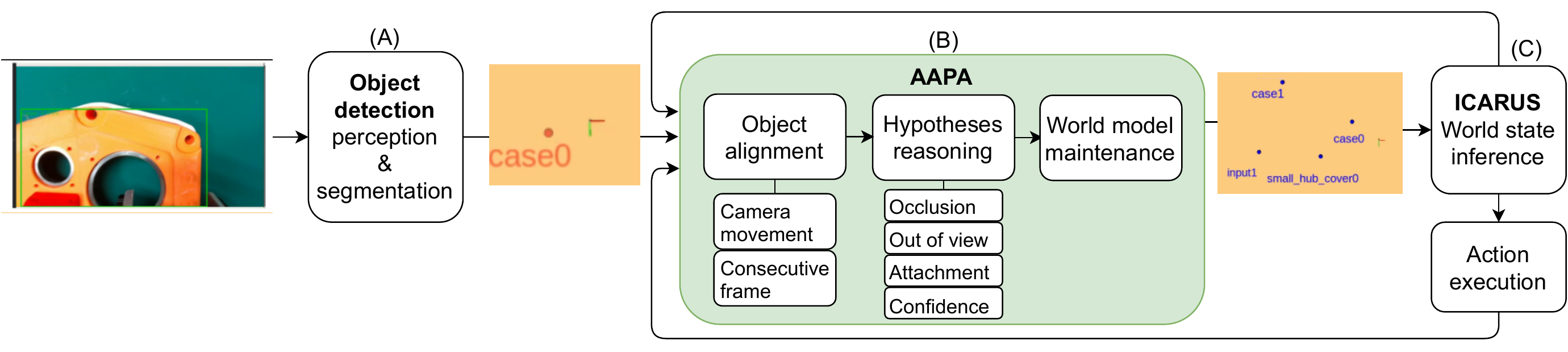}%
 	\vskip -0.15 in
 	\caption{Pipeline of the perceptual anchoring process: A) the object detection module takes the camera input and passes a list of object positions to AAPA, B) which uses this list to maintain anchors in the world model, and C) is subsequently used for inference of world states and relations and action execution. The complete world state and action information are passed to the AAPA in the subsequent cycle.}
	\label{fig:model-overview}
	\vskip -0.15in
\end{figure*}

% Our rule-based model can be used as part of a cognitive architecture.
These assumptions allow us to make reliable inferences for the positions of objects that are no longer detected by the perceptual system but are physically still present in the scene.
The end-to-end complete pipeline consists of three components (Fig.~\ref{fig:model-overview}): A) object detection and segmentation, B) action-aware perceptual anchoring, and C) world state inference and action execution.
In the following, we describe each component in more detail.

\vspace{-0.02in}
\subsection{Object detection and segmentation}
This component performs the initial object detection and feature extraction. It can be replaced by any object detection module as long as it produces a list of object positions and dimensions. The output is a list of percepts $\Pi = \{\pi_1, \pi_2, \dots \}$ each associated with a unique identification label and a set of attributes $\Phi_i = \{\phi_i^{type},\phi_i^{pos},\phi_i^{size}\}$ corresponding to the object category, its position and dimension of the bounding box. $\phi_i$ can also include other attributes such as colour ($\phi_i^{rgb}$) that can be used in the anchoring component.
Since object detection is an on-going research problem, the assigned labels may not always correspond to the same object in the environment, especially if there are multiple objects of the same category. 
This will be handled by the perceptual anchoring model in the following Sec.~\ref{subsec:op-aa}.
In our experiments on the LA-CATER dataset (Sec.~\ref{sec:experiment}) and demonstration with the UR5 (Sec.~\ref{sec:ur5-experiment}) we used a Faster R-CNN (fine-tuned to the respective domains) to output $\Phi_i = \{id,\phi_i^{type}, \phi_i^{pos}, \phi_i^{size}\}$ for each detected object $\pi_i$.

\vspace{-0.011in}
\subsection{Object permanence with action-awareness (AAPA)}
\label{subsec:op-aa}
Our model takes the list of detected objects $\Pi$ and anchors them to the maintained symbol system $\mathcal{P}$.
The anchoring process consists of two main sub-components: object alignment and hypothesis reasoning.
\subsubsection{Object alignment}
The first part of our model addresses the alignment problem, \ie finding the correspondence between objects in two consecutive frames. In our case, we find a mapping between the symbol system $\mathcal{P}_t$ at time $t$ and the new object percepts  $\Pi_{t+1}$ in two steps:
\paragraph{Camera movement alignment}
Whenever the robot's camera moves, the previously maintained object positions will not be correct, especially when dealing with eye-in-hand robotic systems. Thus, we update all object positions in the current world model  $\mathcal{P}_t$ to compensate for the camera movement.
Given the respective camera positions at time $t$ and $t+1$ as $\theta_t$ and $\theta_{t+1}$ in a frame of reference,
%represent the camera's 6 DoF configuration in the camera's frame of reference.
we adjust the position $\phi_i^{pos}$ of an object $p_i \in \mathcal{P}_t$ to $\hat{\phi_i}^{pos} = \phi_i^{pos}+| \hat{\theta}_{t+1}-\hat{\theta}_t|$,
where $\hat{\theta}$ are the camera positions transformed onto the camera frame where $\phi_i^{pos}$ is written.
% \begin{equation}\label{eq:cam-move}
%     \hat{\phi_i}^{pos}(t+1) = |r(t+1) - r(t)|.
% \end{equation}
% This will allow us to maintain object anchors that go out of view due to camera movements, as we assume that state changes are observable 
This also allows us to track objects that go out of view due to camera movements, based on our observability assumption (H1).

\paragraph{Consecutive frame alignment} \label{sssec:munkres}
Using the adjusted positions, we now want to find the optimal alignment between the estimated object states $\hat{\mathcal{P}}_t=\{\hat{p}_j : 1 \dots J\}$ and the newly perceived objects $\Pi_{t+1} = \{\pi_i : 1 \dots I \}$.
We define $C$ as a cost matrix to measure the dissimilarity between $\hat{p}_j$ and $\pi_i$, where each entry $C(\pi_i, \hat{p}_j)$ is determined by
\begin{equation}\label{eq:cost-matrix}
    C(\pi_i,\hat{p}_j) =  \psi * (d^{}(\hat{\phi}_i^{pos},\phi_j^{pos}) + d^{}(\hat{\phi}_i^{size},\phi_j^{size}))
\end{equation}
where $\psi$ denotes a type mismatch constant and $d(\cdot,\cdot)$ a distance measure such as the $L^2$-distance. In our implementation we used the Munkres assignment algorithm~\cite{munkres1957algorithms} to find the optimal alignment, but this can be replaced by any other alignment algorithm.
We create a new symbol system $\mathcal{P}_{t+1} = \{p_k : k = 1 \dots K  \}$ with $K \leq J$ and elements ${p}_j = \hat{p}_j$ for which $C(\pi_i,\hat{p}_j) < \tau$, where $\tau$ is a maximum threshold that depends on the reliability of the object detection output. We update the position and size of ${p}_j$ with $\Phi_i$, the attributes of $\pi_i$ and assign a confidence value $c_i \in [0,1]$ which will be used in the hypothesis reasoning component.
% Since we assume objects changes to be gradual (H2), we only accept assignments where $C(\pi_i,p_j) < maxCost$, a maximum threshold that can be adjusted depending on the reliability of the object detection model. Objects that have not been assigned successfully will be considered in the subsequent reasoning process.
% This algorithm can be replaced with any other alignment algorithm.
% Similarly to~\cite{stiefelhagen2006clear, Li2009Learning}.?
\subsubsection{Hypothesis reasoning} \label{sssec:hypothesis}
The second part of our model considers objects, that have not been assigned successfully in the alignment process. This includes previously maintained anchors $\hat{p}_j \in \hat{\mathcal{P}}_t$ with $\hat{p}_j \notin \mathcal{P}_{t+1}$, as well as newly detected objects $\pi_i \in \Pi_{t+1}$.
Based on our hypotheses on object state changes, we reason about occlusions, potential detection errors and other state changes to decide if their  anchors should be maintained.

\paragraph{Occlusions and out of view}
Objects are considered \textit{occluded} if their maintained anchor $\hat{p}_j \in \hat{\mathcal{P}}_t$ intersects with a detected object. More formally, if there exists $\pi_i \in \Pi_{t+1}$ such that the bounding boxes overlap, \ie $\hat{p}_j \cap \pi_i \neq \emptyset$.
% ($\exists \pi_j \in \Pi: p_i \cap p_j $). %If an object is inserted into another one, it will still be maintained
Objects are considered \textit{out of view} if they have not been assigned a percept and their predicted position $\hat{\phi_j}^{pos}$ is outside of the agent's field of view. %This occurs often after camera movements, especially for eye-in-hand systems.
% Depending on the size of the object, the sensors might not detect them if they are at different points near the margins of the camera view (\eg a big case is not recognised at $x=0.94$, but a smaller plug is still detected at $x=0.98$). We therefore set a wider OOV margin to capture larger objects.
In both cases, the anchor is maintained in the world model and $\hat{p}_j$ is added to $\mathcal{P}_{t+1}$.
If an object is removed from the view without an agent action, it is marked as a \textit{lost} object.

\paragraph{Object attachments}
% Concepts describe aspects of a situation in the environment with a predicate as the head, perceptual matching conditions, tests against matched variables, and references to any sub-relations.
% \paragraph{Definition 2 (Skills}
% Skills describe procedures to achieve certain concept instances in the environment, with a named head, preconditions that need to be true to execute, direct actions to perform in the world or any sub-skills, and the intended effects of the execution.
% These are hierarchical versions of STRIPS operators~\cite{fikes} with a named head, perceptual matching conditions, preconditions that need to be true to execute, direct actions to perform in the world or any sub-skills, and the intended effects of the execution.
We define the semantic relation where an object is \textit{attached} to another object and refer to them as child and parent respectively: if there exists a symbol $p_k \in \mathcal{P}_{t+1}$ and a relation {\tt attached}$(p_j, p_k)$, then the child $p_j$ is considered to be attached to the parent $p_k$ and updated to follow $p_k$.
%$\phi_j^{pos} = \hat{\phi}_j^{pos} + |\phi_k^{pos}(t+1) - \phi_k^{pos}(t)|$ 
% The child moves with the parent until the attachment relation is removed. 
This way, we can track $p_j$, even when it is not detected, as long as its parent is visible. 
This relation is inferred from effects of executed agent actions, \eg 
 for {\tt pick-up}(hand,obj), the world state changes from {\tt is-empty}(hand) to {\tt holding}(hand,obj), from which we can infer {\tt attached}(obj,hand). Similarly for {\tt insert}(obj,case) leading to {\tt inserted}(obj,case) and to {\tt attached}(obj,case). 
We assume that an object can only have one parent but one parent can have multiple children. However, more attachment relations can hold at the same time, allowing for an attachment hierarchy \eg {\tt attached}(plug,case) and {\tt attached}(case,hand). In this case, an object anchor (\eg for \textit{plug}) will be maintained if it has a higher-order parent that has already been anchored (\eg \textit{hand}). In our implementation, we recursively check for attachment relations to maintain anchors for all children. 

\paragraph{Confidence}
Each anchor $p_i\in\mathcal{P}$ is assigned a confidence value $c_i\in[0,1]$.
If the object is first detected, we assign $c_i=0$ and for each subsequent cycle where it is successfully aligned, $c_i$ is increased by a fixed increment \textit{conf}$^+\in (0,1)$, where we cap $c_i\leq1$. If the object disappears, but has previously reached an anchoring threshold $c_i\geq\kappa_{anch}$, it will be considered by the hypothesis reasoning component. Otherwise its confidence value will be decreased by a decrement \textit{conf}$^-\in (0,1)$ and when it reaches $c_i<0$ it will no longer be maintained in the world model.
This allows us to handle false positives or negatives from noisy data, where objects appear or disappear intermittently (H3).

\subsubsection{World model maintenance}
At each cycle we maintain a complete world model, where those that have not been consistently perceived ({$c_i<$ $\kappa_{anch}$}) are not considered for hypothesis reasoning, while the rest ({$c_i\geq$ $\kappa_{anch}$}) are anchored. We can optionally define another confidence threshold \textit{$\kappa_{inf}\geq\kappa_{anch}$} to filter out objects, that we are confident about, to be considered for the world state inference (discussed in Sec.~\ref{subsec:world-state}). These thresholds allow the model to be adjusted to the application scenario that depends on the reliability of the object detection output (\eg higher \textit{$\kappa_{anch}$} to handle flickering more detection errors).

\subsection{World state inference and action execution}
\label{subsec:world-state}
The final component of the pipeline consists of the world state inference. It uses the maintained symbol system $\mathcal{P}_{t+1}$ to infer world states $\mathcal{S}_{t+1}(\supseteq\mathcal{P}_{t+1})$ that additionally include relations between the maintained objects such as spatial relations (\eg {\tt behind}$(snitch,ball)$) or object attachments (\eg {\tt attached}$(snitch,cone)$) inferred from executed actions ({\tt insert}$(snitch,cone)$). The world state $\mathcal{S}_{t+1}$ is then used for task planning to infer the next executable action.
% This step infers object attachments that are caused by previously executed actions (\eg oil plug screwed into case).
It relies on having correctly anchored objects, as an incorrect world model would yield wrong world states and may not find any executable actions.
Finally, the complete world model and executed action are passed as input to the perceptual anchoring model in the subsequent cycle.

%%%%%%%%%%%%%%%%%%%%%%%%%%%%%%%%%%%%%%%%%%%%%%%%%%%%%%%%%%%%%%%%%%%%%%%%%%%%%%%%

\section{Experimental Evaluation}
\label{sec:experiment}
We perform a quantitative evaluation of our model and benchmark against Shamsian \etal \cite{shamsian2020learning} on their LA-CATER (Localisation Annotations) dataset on a snitch localisation task.
We compare model performances when using a Faster R-CNN model as the object detector to using \textit{perfect perception} (ground-truth object locations). 

\subsection{Dataset} 
The LA-CATER dataset~\cite{shamsian2020learning} was generated from the CATER dataset~\cite{girdhar2020cater} which contains synthetically rendered videos (10-second, 300-frame) of standard 3D objects (cube, sphere, cylinder, inverted cone) moving in space. The objects can perform four actions (slide, rotate, pick and place, contain), where `contain' describes a cone moving on top of an object and \textit{containing} it. The dataset contains 14K videos split into train (9.3K), validate (3.3K), and test (1.3K) datasets.
% Each object is characterized by its shape (cube, sphere,cylinder and inverted cone), size (small, medium, large), material (shiny metaland matte rubber) and color (eight colors). 
Every video contains a `snitch', a small golden sphere that is used as the target object for the localisation task.
The LA-CATER dataset augments CATER with frame-level annotations and ground-truth locations for the entire video sequence as well as labels for four object permanence subtasks (visible, occluded, contained, carried). 

\subsection{Baselines} 
We compare our model to the neural model OPNet and a programmed heuristic model both proposed in~\cite{shamsian2020learning} and reran them using the code provided by the authors. %\footnote{Available at {http://github.com/ofrikleinfeld/ObjectPermanence}}.
The heuristic model switches from tracking the target object to tracking the object located closest to the last known location of the target.
% This simple heuristic can be error prone to complex scenarios as it relies entirely on the vision data and can confuse the object to track with one that is merely closeby.
OPNet is a deep network architecture that uses two LSTMs as reasoning modules to infer the object to track (if the target is carried or contained) and its exact location.
OPNet has been trained and validated on the LA-CATER dataset and outperforms other baseline models, including four `learned' models and two `programmed' models, including the heuristic model. As the heuristic model was reported to perform similarly on carried tasks (IoU: OPNet 56.04\% vs Heuristic 55.87\%), we decided to include it in our evaluation.

Since we work with synthetically generated data, the detected objects can be considered more reliable and contain little to no noise. Therefore, we evaluate our AAPA model with $\tau=6500$ for the object alignment threshold, \textit{$\kappa_{anch}=0.1$} and confidence values \textit{conf$^+ =$ conf$^-=0.1$}. This means that objects that appear in two consecutive frames are considered for hypothesis reasoning. For object attachments, we use the action annotations from the LA-CATER dataset and trigger an attachment relation when a `contain' action has been executed.

All models receive bounding box information from the object detector (a Faster R-CNN fine-tuned on the LA-CATER dataset). As misclassified objects can affect the overall performance of the models, we also perform comparisons using \textit{Perfect Perception}, \ie we use the ground-truth locations of the object bounding boxes. This allows us to compare only the failures that occur during reasoning.
% To generate the results for the $L^2$-distance metric, we used code provided by~\cite{shamsian2020learning} and integrated our AAPA model.

\subsection{Evaluation Metrics}
Similar to ~\cite{shamsian2020learning}, we use the mean intersection over union (IoU) of predicted and ground truth bounding boxes as an evaluation metric. However, in order to exclude biases from false predictions, we only compute the metric from the moment, when the snitch is first observed by the object detector, resulting in differing IoU values than reported in~\cite{shamsian2020learning}.
As we are mainly interested in the object location, less in the detected bounding box, performances might differ if predicted bounding box dimensions are wrong. Thus, we also use mean $L^2$-distances (given in pixels) of the centre points of the predicted bounding boxes as a second performance metric. As a reference, the video size is 360x240 pixel, the ground-truth size of the snitch ranges from 15x16 to 41x41 and of any object in the scene from 8x8 to 117x117. 

\begin{table*}[ht]
  \centering
  \caption{Results comparing mean IoU (higher is better) and $L^2$-distances (lower is better)}
   \vskip -0.15in
  \label{tab:opnet-comparison-od}
  \tikzmath{real \r; \r=2.1; }
  \begin{tikzpicture}[scale=1.0]
    \path (\r+.5, 0) node[right, scale=1.0] {%
\begin{tabular}{rccccc}
\hline
\multicolumn{6}{|c|}{\textbf{I. Using Object Detection (OD)}} \\
\hline
Mean IoU $\pm$SEM & Visible & Occluded & Contained & Carried & Overall \\
\hline
Heuristic 	& 88.00 $\pm$0.15  &  59.68 $\pm$0.76  & 55.67 $\pm$0.52  &  51.92 $\pm$0.56  &  	75.70  $\pm$0.40  \\ 
OPNet 	    & \textbf{88.98 $\pm$0.15}  &  80.19 $\pm$0.51  & 77.07 $\pm$0.61  &  56.04 $\pm$0.77  &  	82.35  $\pm$0.39  \\
AAPA	& 88.67 $\pm$0.16  &  \textbf{82.15 $\pm$0.41}  & \textbf{80.79 $\pm$0.34}  &  \textbf{68.25 $\pm$0.43}  &  	\textbf{84.66  $\pm$0.23}  \\
% \hline\hline
% Mean $L^2$-distance $\pm$SEM & Visible & Occluded & Contained & Carried & Overall \\
% \hline
% Heuristic & \textbf{1.47 $\pm$0.04} & 10.04 $\pm$0.59 & 10.13 $\pm$0.53 & 11.44 $\pm$0.95 & 5.36 $\pm$0.29 \\
% OPNet & 1.50 $\pm$0.11 & \textbf{4.61 $\pm$0.49} & 8.10 $\pm$0.58 & 14.36 $\pm$0.95 & 5.10 $\pm$0.33 \\
% AAPA (mW = 6500) & 1.67 $\pm$0.19 & 6.27 $\pm$1.11 & \textbf{3.05 $\pm$0.47} & \textbf{4.67 $\pm$0.37} & \textbf{3.09 $\pm$0.44} \\
\hline\hline
Mean $L^2$ $\pm$SEM & Visible & Occluded & Contained & Carried & Overall \\ % originally called L2-cropped
\hline
Heuristic & 1.43 $\pm$0.02 & 8.69 $\pm$0.31 & 9.41 $\pm$0.39 & 10.98 $\pm$0.50 & 4.79 $\pm$0.22 \\
OPNet     & \textbf{1.42 $\pm$0.07} & 3.44 $\pm$0.23 & 7.33 $\pm$0.41 & 13.97 $\pm$0.51 & 4.53 $\pm$0.26 \\
AAPA     & 1.44 $\pm$0.09 & \textbf{2.44 $\pm$0.18} & \textbf{2.25 $\pm$0.13} & \textbf{4.65 $\pm$0.21} & \textbf{1.96 $\pm$0.11} \\
\hline
& & & & & \\
\hline
\multicolumn{6}{|c|}{\textbf{II. Using Perfect Perception (PP)}} \\
%  & &\textbf{Perfect} &\textbf{Perception}& &\\
\hline
Mean IoU $\pm$SEM & Visible & Occluded & Contained & Carried & Overall \\
\hline
Heuristic 	& 96.83 $\pm$0.23  &	36.32 $\pm$0.89  & 	60.66 $\pm$0.46  & 	59.87 $\pm$0.49  & 	82.86 $\pm$0.43\\ 
OPNet 	    & 90.02 $\pm$0.11  &	73.81 $\pm$0.57  & 	84.40 $\pm$0.42  & 	77.08 $\pm$0.64  & 	87.09 $\pm$0.28\\
AAPA	    & \textbf{99.38 $\pm$0.12}  &	\textbf{90.53 $\pm$0.64}  & 	\textbf{93.86 $\pm$0.33}  & 	\textbf{82.54 $\pm$0.34}  & 	\textbf{96.31 $\pm$0.22}\\

% \hline\hline
% Mean $L^2$-distance $\pm$SEM & Visible & Occluded & Contained & Carried & Overall \\
% \hline
% Heuristic & 1.94 $\pm$0.23 & 20.32 $\pm$0.83 & 7.15 $\pm$0.35 & 7.57 $\pm$0.66 & 4.65 $\pm$0.27 \\
% OPNet & \textbf{1.72 $\pm$0.14} & \textbf{7.44 $\pm$0.54} & 4.17 $\pm$0.41 & 6.30 $\pm$0.66 & \textbf{3.32 $\pm$0.27} \\
% AAPA (mW = 6500) & 2.65 $\pm$0.45 & 14.66 $\pm$1.76 & \textbf{2.88 $\pm$0.69} & \textbf{3.21 $\pm$0.77 }& 4.10 $\pm$0.62 \\
\hline\hline
Mean $L^2$ $\pm$SEM & Visible & Occluded & Contained & Carried & Overall \\ % originally called L2-cropped
\hline
Heuristic & 0.98 $\pm$0.10 & 15.53 $\pm$0.35 &  6.52 $\pm$0.24 & 7.16 $\pm$0.34 &  3.43 $\pm$0.16 \\
OPNet     & 1.07 $\pm$0.04 & 4.16 $\pm$0.17 &  3.44 $\pm$0.28 & 5.75 $\pm$0.34 & 2.28 $\pm$0.17 \\
AAPA     & \textbf{0.28 $\pm$0.08} & \textbf{2.51 $\pm$0.23} &  \textbf{0.77 $\pm$0.15} & \textbf{1.94 $\pm$0.17} & \textbf{0.62 $\pm$0.11} \\
\hline
\end{tabular}
    };
    % \fill[green!30!black] (0, 0) circle (\r);
    % \fill[gray!40] (0, 0) -- (\r, 0) arc (0:90:\r) -- cycle;
     
    \path (\r+.6, 0) node[left, scale=1.0] 
    %{%
    % \path (\r*.45, \r*.2)
    % node[text width=8em, align=left, scale=.7]
    {\includegraphics[width=20.5em]{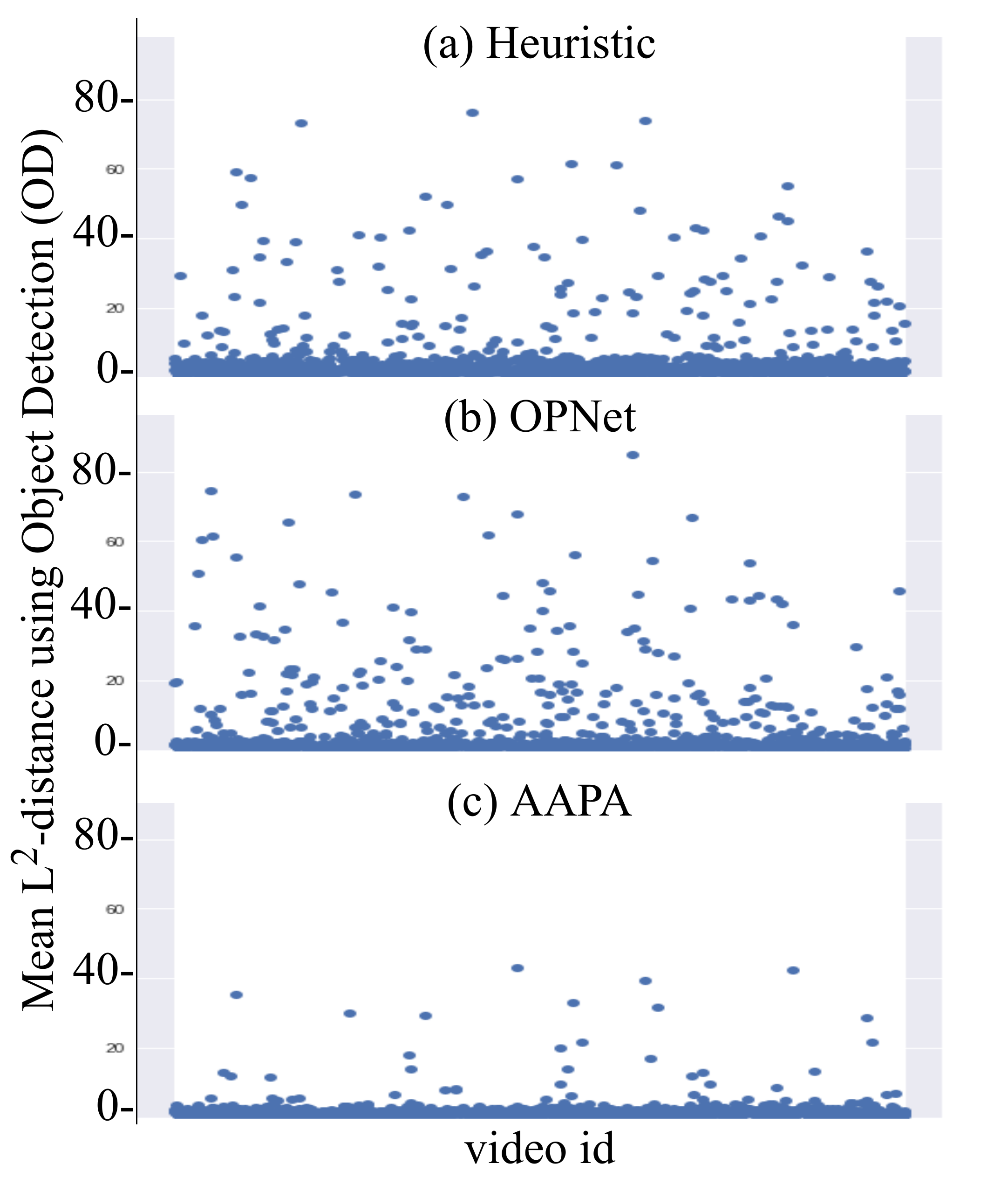}};
    % \path (-\r*.45, -\r*.4)
    % node[white, text width=6em, align=center, scale=.7] {\textbf{Stochastic 75\%}};
  \end{tikzpicture}
    	\vskip -0.2in
\end{table*}
\vskip -0.2in

% \begin{figure}[thpb]
% 	\centering	
% 	\vskip -0.1in
%  	\includegraphics[width=1\linewidth]{figures/l2-chart-od.pdf}%
%  	\vskip -0.1in
%  	\caption{Distribution of $L^2$-distances for the 1,371 videos predicted by the different models using OD bounding boxes.}
% 	\label{fig:l2-distribution}
%  	\vskip -0.1in
% \end{figure}

\subsection{Results}
In the following we present results using object detection (OD) and perfect perception (PP) and both mean IoU and L$^2$-distances.
We compare performances on different task categories (visible, occluded, contained, carried) and discuss common failure cases and their causes.

\subsubsection{Using object detection (OD)}
Table~\ref{tab:opnet-comparison-od}.I shows a summary of the model performances comparing AAPA with OPNet and Heuristic by mean IoU and $L^2$-distance. %\footnote{Examples of model performances can be seen in the attached video.}
AAPA outperforms the baseline models in all categories, except on visible tasks, where it performs similarly (AAPA: 88.67\% vs OPNet: 88.98\% vs Heuristic: 88.00\%). 
AAPA's performance drops for occluded tasks, mainly due to the snitch moving in occlusion as we assume that objects do not move on their own (H1). Other errors included incorrect prediction of bounding box size for partial occlusions.
Similarly, for contained tasks, most of the failures occurred when containment happened after the snitch was moving in occlusion. %But there are fewer bounding box errors, as we retain the last detected size.
For carried tasks, AAPA performs a lot better than the other models  (AAPA 68.25\% vs OPNet's 56.04\% vs Heuristic 51.92\%). Main failures include moving in occlusion (while being carried) and anchoring errors caused by the threshold parameter $\tau$ in the object alignment. The mean IoU performance is also impacted by the fact that AAPA retains the last detected bounding box size before it became contained, while the ground truth size becomes smaller or larger, when the object is carried in 3D space.
% As AAPA does not focus on predicting the bounding box size but only the location, it does not adjust the size when objects move within the 3D space, resulting in less accurate IoU.
While OPNet adjusts the bounding box size, when the snitch is carried (which it learned during training), it still underperforms AAPA for carried tasks.
% We further consider the parameter \textit{$\kappa_{anch}$} to set the number of cycles for which an object needs to be present in order to be considered for anchoring. This is required to avoid anchoring false positive objects that only appear for one cycle. This parameter needs to be adjusted depending on the reliability of the detector and if we deal with simulated or real data.

\subsubsection{Using perfect perception (PP)}
When using ground truth bounding boxes instead of the object detector, no detection errors are propagated to the models, allowing us to evaluate their reasoning capabilities.
All model performances improved, with AAPA improving the most on carried tasks, outperforming the baseline models in all tasks for both mean IoU and $L^2$-distance (Table~\ref{tab:opnet-comparison-od}.II).
For visible tasks, AAPA and Heuristic benefit the most from ground truth information, as there is no bounding box alignment error.
% As OPNet learned to adjust the bounding box position and size during the tracking process, these minor adjustments decreased its overall performance, so it did not improve on visible tasks.
The main failure cases are caused by moving in occlusion and anchoring errors. The impact of AAPA not adjusting the bounding box size while being carried in 3D space can be seen when comparing mean IoU to $L^2$-distances: the mean IoU for AAPA is better for occluded tasks (90.53\%) compared to carried tasks (82.54\%), but the mean $L^2$-distance shows the opposite: 2.51 for occluded vs 1.94 for carried (lower is better).

\subsubsection{Discussions}
While many deep learning approaches exist that address object permanence, they require large amounts of data and remain a black-box solution. This means that success and failure cases are not necessarily explainable and there is little room to investigate wrong reasoning processes.
One of the main benefits of our approach is the transparency of our rule-based model, which makes it human readable. This allows us to investigate failure cases and to fine-tune our model parameters \eg that anchoring failures occurred when objects moved in occlusion.
Other failures were due to objects taking a bigger leap between two consecutive frames: Our model considered them as two different object instances, due to our assumption that object positions stay the same as last time they were seen (H1) and they do not `teleport' (H2). If we increase $\tau$, the threshold for consecutive frame alignment, these cases are anchored correctly, but would give a bigger leeway to detection errors, resulting in objects anchored to ghost objects. We experimented with these values and chose a threshold that yielded the maximum overall performance for this task. %OPNet assumes that all objects are unique and that there are no objects of the same type. %It would be interesting to compare OPNet's performance on a dataset with multiple objects of the same type.

The figures next to Table~\ref{tab:opnet-comparison-od} show the distribution of mean $L^2$-distances for all 1,371 videos for Heuristic, OPNet, and AAPA using OD.
AAPA produces more accurate predictions and with only a few outliers, representing the failures mentioned above.
%CLEVR_new_001982
The biggest outlier for AAPA comes from a video, where the snitch was occluded in all frames but one. The baseline models (OPNet and Heuristic) will always produce a prediction for each video frame, even when the snitch is not visible at all.  On the other hand, our AAPA model assumes that object state changes are consistent, and only maintains anchors for objects that are detected in multiple consecutive frames. Therefore, in this case, AAPA fails to maintain the anchor for the snitch and returns no prediction of its position, which the evaluation considers as a $[0,0]$ vector and returns an $L^2$-distance of $|\phi^{pos}|$ given by the ground-truth position.

% \begin{figure}[tb]
% 	\centering	
%  	\includegraphics[width=0.98\linewidth]{figures/chart-baseline-iou-by-frames.png}%
%  	\caption{Model performances comparing mean IoU by number of frames where the snitch is contained by multiple cones (hierarchical containment), with perfect perception (pp) and Faster R-CNN object detection (od). Our model AAPA outperforms OPNet and the Heuristic model for both}
% 	\label{fig:chart-comparison}
% \end{figure}
\vspace{-0.02in}
\section{Real-world application}
\label{sec:ur5-experiment}
We evaluate our model qualitatively and demonstrate its use in real-world applications on a collaborative gearbox assembly task with the Universal Robot UR5, an industrial robotic arm with 6DoF and 5kg payload.
The robot is mounted with an Intel RealSense D435 camera, acting as an eye-in-hand system (Fig. \ref{fig:setup}).
AAPA has been implemented as part of ICARUS, a cognitive architecture for modelling human cognition~\cite{langley2006unified}, which uses the world model for world state inference and task execution (Fig.~\ref{fig:model-overview}C).
For the object detection (Fig.~\ref{fig:model-overview}A) we use a Faster R-CNN that has been fine-tuned to object categories in the gearbox assembly domain (\eg casing top/base, input subassembly). 
We found the optimal confidence parameters for this scenario to be $\kappa_{anch} = 0.5, \kappa_{inf} = 0.8,$ \textit{conf$^+ = 0.05$} and \textit{conf$^- = 0.1$}.\footnote{Full task execution can be seen on \text{https://youtu.be/3qCkzQS9xk4}} 
%Since As AAPA only needs a list of objects with their detected positions and dimensions, this detection model can be replaced with any other model. 

\begin{figure*}[ht]
	\centering	
 	\includegraphics[width=0.95\linewidth]{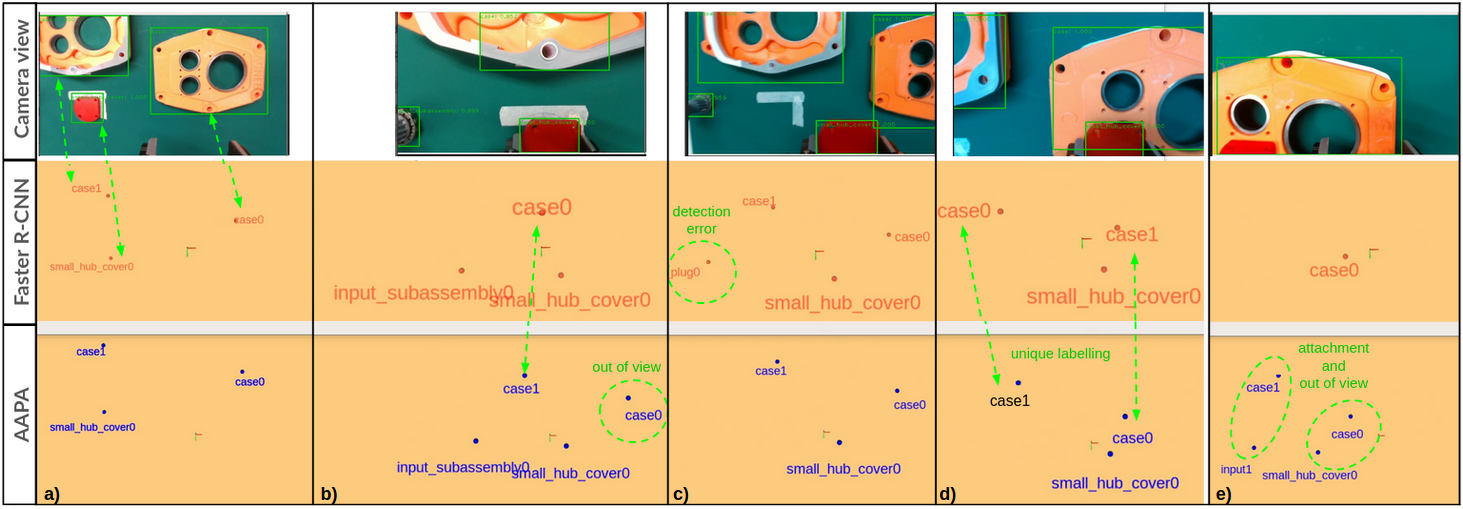}%
 	\vskip -0.1in
 	\caption{Gearbox assembly: eye-in-hand camera view (top), detected objects using Faster R-CNN (middle) and anchored world model using AAPA (bottom). AAPA tracks objects that go out of view (b,e), handles detection errors (c), maintains correct unique labels (b,d) and objects attachments (e).}
	\vskip -0.25in
	\label{fig:demo-screencaps}
\end{figure*}

\subsection{Gearbox assembly scenario}
\vskip -0.05in
For autonomous navigation and intelligent collaboration, robots must track objects, even if they are no longer perceived by the detection system. For eye-in-hand robotic systems, this is particularly important as the grasping action generally results in losing sight of most surrounding objects.
We demonstrate the use of AAPA on an assembly task, where the robot has to manipulate gearbox assembly parts in collaboration with a human operator.
The cognitive architecture uses the world model maintained by AAPA to reason about the necessary actions to execute. %The robot recognises attachments that result in inserting or attaching objects.
The collaborative gearbox assembly consists of the following steps:
\begin{enumerate}
    \item Robot attaches small hub cover onto casing top (case0).
    \item Human screws in hub cover plugs.
    \item Robot inserts input subassembly to casing base (case1).
    \item Human moves casing base to the robot.
    \item Robot mounts casing top to casing base.
\end{enumerate}

% \begin{figure}[thp]
% 	\centering
%  	\includegraphics[width=0.95\linewidth]{figures/rosbag_pickplaceshaf\kappa_tracking_141.jpg}
%  	\caption{Object detection using the Faster R-CNN fine-tuned on our assembly task.}
% 	\label{fig:fcnn}
% \end{figure}

% \begin{figure}[h]
% 	\centering
%  	\includegraphics[width=0.95\linewidth]{figures/plug_case_tracking_5.jpg}
%  	\caption{Object detection using the Faster R-CNN of the assembled case with hub covers.}
% 	\label{fig:case-hubcovers}
% \end{figure}

\subsection{Observations and results}
\vskip -0.05in
The robot successfully executed the gearbox assembly task in real-time (see Fig.~\ref{fig:demo-screencaps}).
% In particular for assembly tasks, the individual parts become partially or completely occluded throughout the assembly task. 
We compare the maintained world state with and without our AAPA model.

\subsubsection{Unique label assignments}
The Faster R-CNN detects objects and assigns unique labels in the format \textit{typeN}, where $N$ is the instance number assigned in order of appearance (\eg case0, case1).
When there are multiple objects of the same type, it can produce incorrect labels as the order of appearance changes or an object disappears from the view. AAPA maintains the unique labels of the anchors with the consecutive frame alignment (Fig.~\ref{fig:demo-screencaps}b,d). %(results/rosbag pickplaceshaft tracking: casing bolts get mixed up) 
% or when the objects go out of view or move too quickly 
%(results/shafts tracking: input shaft gets renamed from 12.1 to 12.4, results/ plugs tracking, plugs shaft tracking, plug case tracking: small hub covers)
\subsubsection{Occlusion and out of view}
Once objects are occluded or go out of view, the Faster R-CNN loses track of them, which can lead to failures in the world state inference and action execution. With AAPA, the anchors of these objects are maintained in the world model (Fig.~\ref{fig:demo-screencaps}b,e).

\subsubsection{Handling detection errors}
When objects are cluttered, stacked, or partially occluded, it sometimes detects the wrong object type or fails to detect them.
State-of-the-art object detection models can be very error-prone due to environmental changes such as lighting conditions, or simply when they are viewed from different angles or in close-up when grasping. These models commonly produce flickering ghost objects that appear and disappear within a couple of frames. The robot needs to handle these detection errors correctly, otherwise it can result in an incorrect world state inference and task execution. Our model handles these errors using the confidence value that is updated at each cycle (Fig.~\ref{fig:demo-screencaps}c).

\subsubsection{Attachment relations}
The assembly task triggers object attachments (\ie attached(small\_hub\_cover0, case0), attached(input\_subassembly0, case1)). When the robot picks up the casing top (Step 5), it fails to detect the surrounding objects, including the hub cover. However, since it is attached to the casing top, its anchor is maintained and updated to follow the casing top as it moves (Fig.~\ref{fig:demo-screencaps}e). Similarly, when the robot grasps objects during task execution, it sometimes fails to detect them. Since the pick-up action creates an attachment between the gripper and grasped object, the anchor is maintained and updated consistently.

%%%%%%%%%%%%%%%%%%%%%%%%%%%%%%%%%%%%%%%%%%%%%%%%%%%%%%%%%%%%%%%%%%%%%%%%%%%%%%%%
% \section{Discussions}
% \label{sec:discussions}
% High-level representations are important for real-world applications, especially for robots interacting with humans, as the decisions making process needs to be explainable~\cite{camilleri2017analysing}. 
\vskip -0.1in
\section{Conclusions}
\label{sec:conclusion}
\vskip -0.05in
We presented an action-aware perceptual anchoring model (AAPA) that can handle object detection errors, out of view, and more complex scenarios such as invisible displacements by considering agent actions.
We showed that AAPA outperforms baseline models on the LA-CATER dataset for a snitch localisation task.
We demonstrated its use for robotic applications on a real-life gearbox assembly task between a human operator and a UR5. AAPA successfully maintains a reliable world model for the robot to complete the task.
% Cognitive development of object permanence in robots enables the development of more intelligent search behaviour. 
% By using high-level reasoning, the focus is shifted away from relying mainly on visual perception to allowing other perceptual modalities \ie our model could perform equally for object locations that come from other sources such as verbal or haptic inputs.
Our approach can be used for hierarchical attachments as well as other agent actions that cause attachments (\eg screwing a plug onto a case).
% One of the main failure cases in the LA-CATER benchmarking was caused by occluded objects in motion, which can be solved using motion prediction but is left for future work. 
% Our current approach requires us to provide a list of attachment relations that can be inferred from agent actions. 
Given a list of agent actions that cause attachment relations, our model recursively checks for these relations and automatically infers the attachment hierarchy irrespective of the number of relations defined.
% As we have not integrated human action recognition, we infer attachments in the collaborative task from existing object states (\eg hub cover inserted to case). 
In the future we will focus on using human action recognition in combination with (visual) commonsense reasoning to infer attachment relations automatically from agent actions and object types (\eg screw action, container objects).

%%%%%%%%%%%%%%%%%%%%%%%%%%%%%%%%%%%%%%%%%%%%%%%%%%%%%%%%%%%%%%%%%%%%%%%%%%%%%%%%
\vskip -0.08in
\small{
\section*{Acknowledgements}
\vskip -0.03in
This research is supported by A$^{*}$STAR under its Human-Robot Collaborative AI for Advanced Manufacturing and Engineering (AME) programme (Grant number A18A2b0046).
}
% \vspace{-0.03in}
% \small{
\bibliography{root.bib}
\bibliographystyle{IEEEtran}
% }

\end{document}